\documentclass{article}

\usepackage[nolist]{acronym}
\usepackage{authblk}
\usepackage{booktabs}
\usepackage{graphicx}
\usepackage{hyperref}
\usepackage{lscape}
\usepackage{natbib}
\usepackage{siunitx}
\usepackage[disable]{todonotes}
\usepackage{multirow}
\usepackage{lineno}

\setcitestyle{authoryear,open={(},close={)}}

\usepackage{amsmath,amssymb,amsfonts,textcomp}
\newcommand{\R}{\mathbb{R}}

\usepackage{mathtools}

\begin{acronym}[]
    \acro{AI}{Artificial Intelligence}
    \acro{ANN}{Artificial Neural Network}
    \acro{BPTT}{Backpropagation Through Time}
    \acro{CA}{Cellular Automaton}
    \acrodefplural{CA}{Cellular Automata}
    \acro{CNN}{Convolutional Neural Network}
    \acro{CT}{Computed Tomography}
    \acro{DoF}{Degrees of Freedom}
    \acro{DSP}{Digital Signal Processing}
    \acro{FPGA}{Field Programmable Gate Array}
    \acro{FPU}{Floating Point Unit}
    \acro{GAN}{Generative Adversarial Network}
    \acro{GI}{Gastro-Intestinal}
    \acro{GNN}{Graph Neural Network}
    \acro{GPU}{Graphics Processing Unit}
    \acro{ID}{In-Distribution}
    \acro{LM}{Levenberg Marquardt}
    \acro{MCU}{Microcontroller Unit}
    \acro{MLP}{Multi-Layer Perceptron}
    \acro{MSE}{Mean-Squared Error}
    \acro{MRI}{Magnetic Resonance Imaging}
    \acro{NQM}{NCA Quality Metric}
    \acro{OCT}{Optical Coherence Tomography\textbf{}}
    \acro{OOD}{Out-of-Distribution}
    \acro{OR}{Operating Room}
    \acro{OTS}{Optical Tracking System}
    \acro{NCA}{Neural Cellular Automaton}
    \acrodefplural{NCA}{Neural Cellular Automata}
    \acro{PDE}{Partial Differential Equation}
    \acrodefplural{PDE}{Partial Differential Equations}
    \acro{PMT}{Permanent Magnet Tracking}
    \acro{PnO}{Position and Orientation}
    \acro{PSRAM}{Pseudo-Static Random Access Memory}
    \acro{RAM}{Random Access Memory}
    \acro{ReLU}{Rectified Linear Unit}
    \acro{SIMD}{Single Instruction Multiple Data}
    \acro{SLAM}{Simultaneous Localization and Mapping}
    \acro{SPI}{Serial Peripheral Interface}
    \acro{VAE}{Variational Autoencoder}
    \acro{ViT}{Vision Transformer}
    \acro{VO}{Visual Odometry}
    \acro{VRAM}{Video Random Access Memory}
    \acro{WCE}{Wireless Capsule Endoscopy}
    \acro{WGAN}{Wasserstein GAN}
    \acro{MRI}{Magnetic Resonance Image}
    \acro{CLA}{Cellular Learning Automaton}
    \acro{ANN-CA}{Artificial Neural Network Cellular Automaton}
    \acro{Markov-CA}{Markov Cellular Automaton}
    \acro{LULC}{Land Use and Land Cover}
    \acro{ANN-CA-MC}{Artificial Neural Network CA with Markov Chain}
    \acro{ACAN}{Asynchronous Cellular Automaton-based Neuron}
    \acro{ANCAN}{Asynchronous recurrent Network of CA-based Neurons}
    \acro{LSTM-CA}{Long Short-Term Memory Cellular Automaton}
    \acro{LSTM}{Long Short-Term Memory}
    \acro{ConvLSTM}{Convolutional Long Short-Term Memory}
    \acro{PCB}{Printed Circuit Board}
    \acro{DLGN}{Differentiable Logic Gate Network}
    \acro{FPGA}{Field-Programmable Gate Array}
    \acro{ASIC}{Application-Specific Integrated Circuit}
    \acro{PNN}{Photonic Neural Network}
    \acro{PRISMA}{Preferred Reporting Items for Systematic reviews and Meta-Analyses}
\end{acronym}

\begin{document}

\title{Applications of Neural Cellular Automata:\\State of the Art, Challenges and Opportunities}

\author[1,2]{Nick Lemke}
\author[1]{Niklas Ihm}
\author[1,3]{John Kalkhof}
\author[1,4]{Mirko Konstantin}
\author[1]{Henry J. Krumb}
\author[5,6]{Daniel M. Lang}
\author[8]{Ehsan Pajouheshagar}
\author[5,6,7]{Ario Sadafi}
\author[1,9]{Arjan Kuijper}
\author[10]{Karim Lekadir}
\author[3]{Marco Lorenzi}
\author[5,6,7,11,12]{Carsten Marr}
\author[5,6]{Julia A. Schnabel}
\author[1]{Anirban Mukhopadhyay}

\affil[1]{Technical University of Darmstadt, Darmstadt, Germany}
\affil[2]{ImFusion GmbH, Munich, Germany}
\affil[3]{Inria Center at University Côte d’Azur, Sophia Antipolis, France}
\affil[4]{Zuse Institute Berlin, Berlin, Germany}
\affil[5]{Helmholtz Munich, Munich, Germany}
\affil[6]{Technical University of Munich, Munich, Germany}
\affil[7]{Munich Center for Machine Learning (MCML), Munich, Germany}
\affil[8]{EPFL, Lausanne, Switzerland}
\affil[9]{Fraunhofer IGD, Darmstadt, Germany}
\affil[10]{Universitat de Barcelona, Barcelona, Spain}
\affil[11]{Ludwig Maximilian University Munich, Munich, Germany}
\affil[12]{Ludwig Maximilian University Hospital, Munich, Germany}

\maketitle

\begin{abstract}
\noindent\acp{NCA} are a new type of neural network architecture which enable accurate and robust inference at extremely small model sizes.
Recently, \acp{NCA} have advanced to become interesting low-resource alternatives to convolution- and attention-based architectures for various tasks such as image analysis, synthetic image generation, and simulation.
The rapid development and increased research interest necessitate a comprehensive review of the emerging technology.
This review provides an overview of the fundamentals of \acp{NCA}, applications to medical imaging, as well as insights into the state of the art.
We analyze recent modifications to the originally proposed \ac{NCA} architecture with respect to their efficiency and accuracy.
Furthermore, we review practical applications in real-world scenarios with a focus on medical image analysis, segmentation, classification, registration, depth estimation, and image synthesis.
Finally, we identify several advantages of \acp{NCA}, research gaps, and conclude with an analysis of future opportunities for \acp{NCA} in medical applications in confined settings or areas that have particular demands for robustness or efficient data processing.

\acresetall
\end{abstract}


\section{Introduction}\label{sec:intro}

With the rise of \ac{AI} in medical image analysis, the research interest in robust deep learning models increases.
Modern \ac{ANN} models have achieved remarkable performance, but complex architectures often come at the cost of increasingly large model sizes, which prohibit their application.
Applications in resource-constrained environments, such as edge devices, therefore demand small and robust solutions.

Recently, \acp{NCA} were proposed as a new type of lightweight self-organizing neural network models that can be employed in various image analysis tasks with in resource-constrained settings with special demands to robustness.
Researchers in the medical imaging community acknowledge the potential of \acp{NCA} being used on small-scale hardware in spatially confined settings, and their robustness against population or manifestation shifts.\citep{ali2024breasta,korevaar2024generalization,kalkhof2025mednca,krumb2025encapsulatea}, classification \cite{deutges2024neurala}, registration \citep{ranem2024ncamorph} or depth estimation \citep{krumb2025encapsulatea}.

Opposed to CNN- or attention-based models, \acp{NCA} rely on emergent properties of \acp{CA}.
\acp{NCA} efficiently trade model size for recurrence (i.e. sequential processing), while often being on par with state-of-the-art models in terms of accuracy~\citep{kalkhof2023mednca}.
Further, \acp{NCA} were shown to be invariant to image size~\citep{kalkhof2023m3dnca}, more robust to certain types of data drift~\citep{kalkhof2023mednca}, and easier to deploy on edge devices without loss of accuracy~\citep{krumb2025encapsulatea}.
All of these properties make \acp{NCA} an attractive choice for applications with practical constraints, such as limited memory or increased chance of data drift.

However, despite all the potential benefits, researchers training \acp{NCA} for tasks in medical imaging will still face several challenges associated with this relatively young model architecture.
For instance, strategies to reduce computational cost at training time are still being investigated and mainly leverage domain-specific rather than general-purpose optimizations.
Further, a full exploration of best practices is still subject to future research.
A systematic search of suitable practices for \ac{NCA} model design, application to more downstream tasks (image detection, compression, and others) or data types (sequential data) is still improvable.

\begin{figure}[ht]
    \centering
    \includegraphics[width=0.8\linewidth]{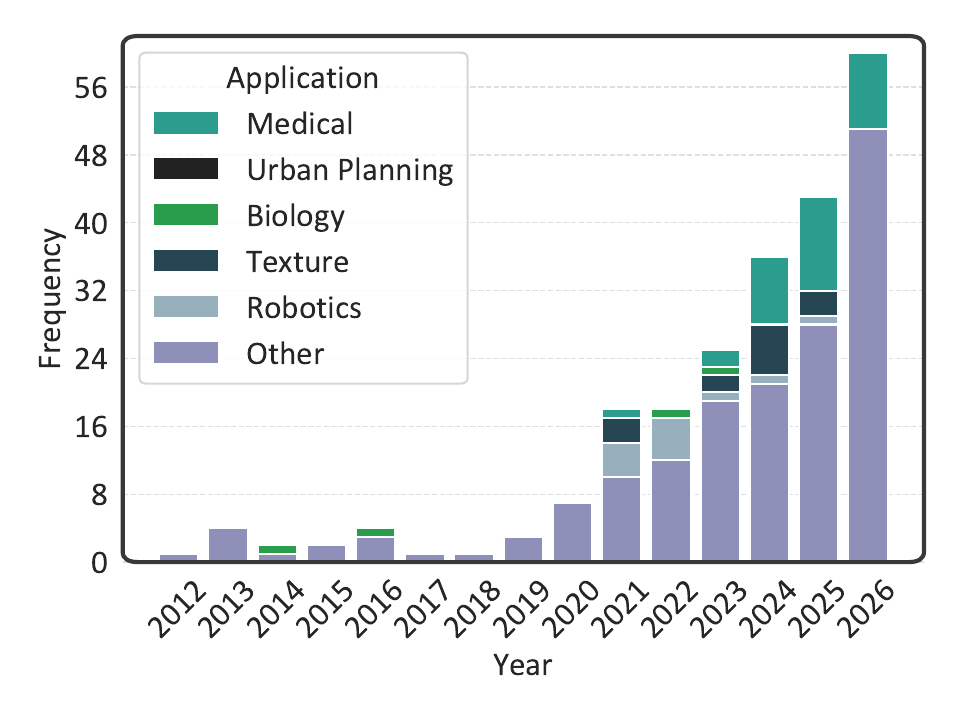}
    \caption{Papers published on \acp{NCA} per year, categorized by application.  
    A steady linear growth of \ac{NCA} papers can be observed since 2020.
    \textbf{}
    }
    \label{fig:papersperyear}
\end{figure}

The increased research interest in \acp{NCA} over the past six years by both practitioners and theorists  reflects in a rising number of papers being published on the topic (Figure~\ref{fig:papersperyear}).
Here, we present the first systematic survey of the recent state of the art in the development of \acp{NCA}, with a focus on their applications in medical imaging.
While \citet{hartl2026neural} have recently published a review article on \acp{NCA}, their literature review focuses on understanding biological pattern formation and self-organization with \acp{NCA}.
In this paper, we present a comprehensive review of \textit{applications} of \acp{NCA} and identify research gaps that are relevant for their practical deployment.

We first outline our method for systematic literature research, adhering to \ac{PRISMA} guidelines, in Section~\ref{sec:method}.
Further, we describe the fundamental concepts behind \acp{NCA} to give the reader an idea of the basic architecture, its hyperparameters, possible tweaks and modifications of the architecture in Section~\ref{sec:fundamentals}.
On that basis, Section~\ref{sec:applications} introduces a variety of possible applications for which \acp{NCA} are employed, what challenges were faced, and how they were overcome.
Finally, we discuss the value of \acp{NCA} for medical imaging and potential opportunities for future research  (Section~\ref{sec:discussion}).

The key findings of this survey include:
\begin{itemize}
    \item Overall research interest in \acp{NCA} increases linearly since their formalization by~\cite{mordvintsev2020growinga}.
    \item We observe a tendency of \ac{NCA} research heading towards medical imaging applications, due to their versatility and small scale.
    Medical image analysis is a driving force for \ac{NCA} developments, as potential tasks, datasets and challenges are exceptionally diverse within this regime.
    \item We identify and exchange ideas from different isolated research streams working on \acp{NCA}, which are usually overlooked as they use different terminology.
    \item We identify several research gaps in \textit{conception}, \textit{assessment} and \textit{deployment} of \acp{NCA} for medical image analysis.
\end{itemize}

\begin{figure}[ht]
    \centering
    \includegraphics[width=0.8\linewidth]{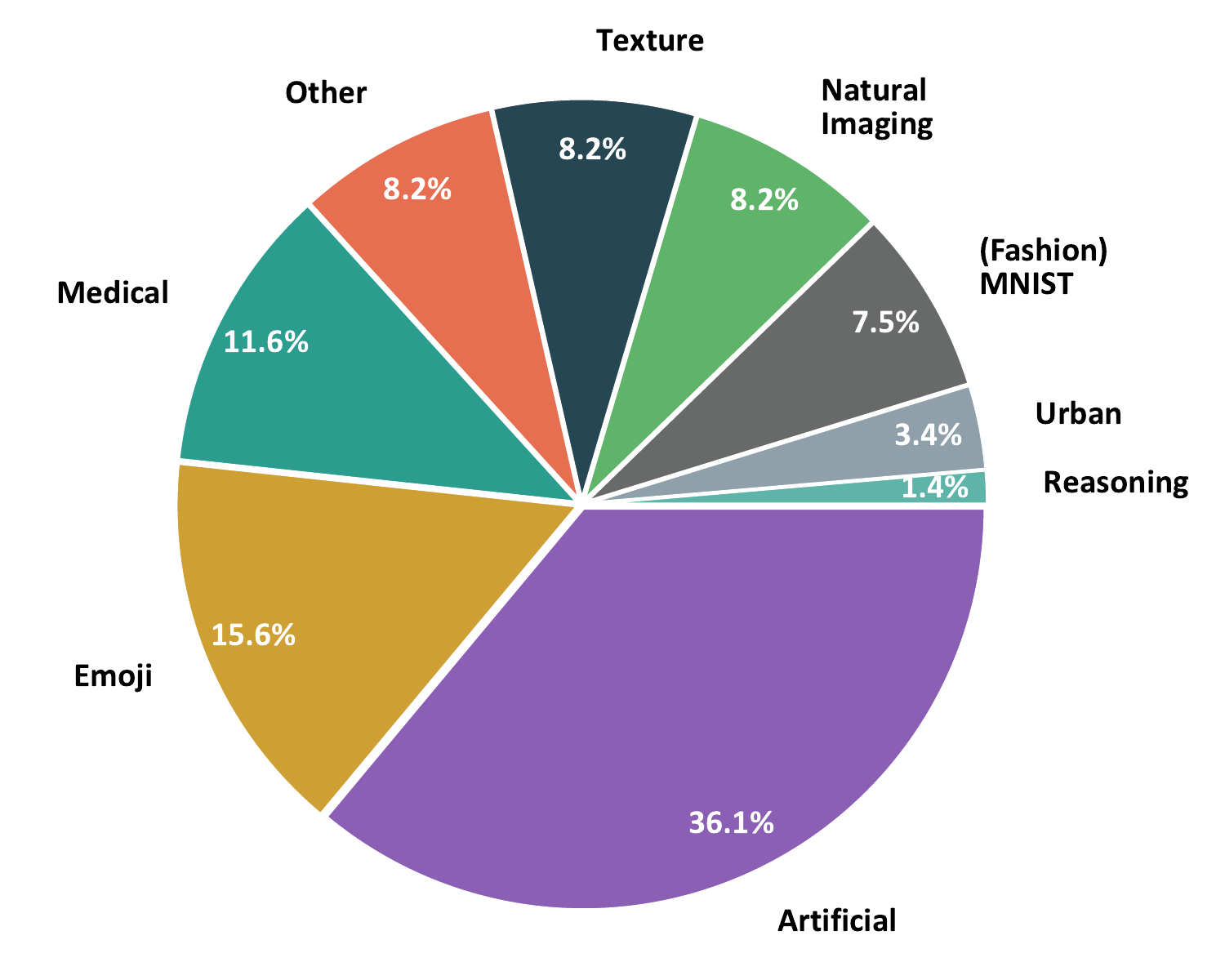}
    \caption{Representation of datasets in \ac{NCA} literature.}
    \label{fig:piedatasets}
\end{figure}

\section{Method of Literature Research}\label{sec:method}

For our literature review, we query Pubmed, Google Scholar, and Semantic Scholar for the search term \textit{"Neural Cellular Automata"} (in quotes).
The cut-off point for this search was 31 August 2026.
PubMed yielded 178 entries, of which 171 were considered relevant.
Google Scholar yielded 559 results, of which 149 were considered relevant, whereas Semantic Scholar yielded 107 relevant entries out of 109 entries total.
Apart from a systematic literature review, we further added 7 papers to the pool that we learned about through conferences and conversations with peers.

Initially, we removed duplicate entries across search indices and those that did not meet the criteria of research papers, including unpublished abstracts, retracted papers, and papers written in languages other than English, yielding 396 relevant entries total.

\begin{figure}
    \centering
    \includegraphics[width=0.8\linewidth]{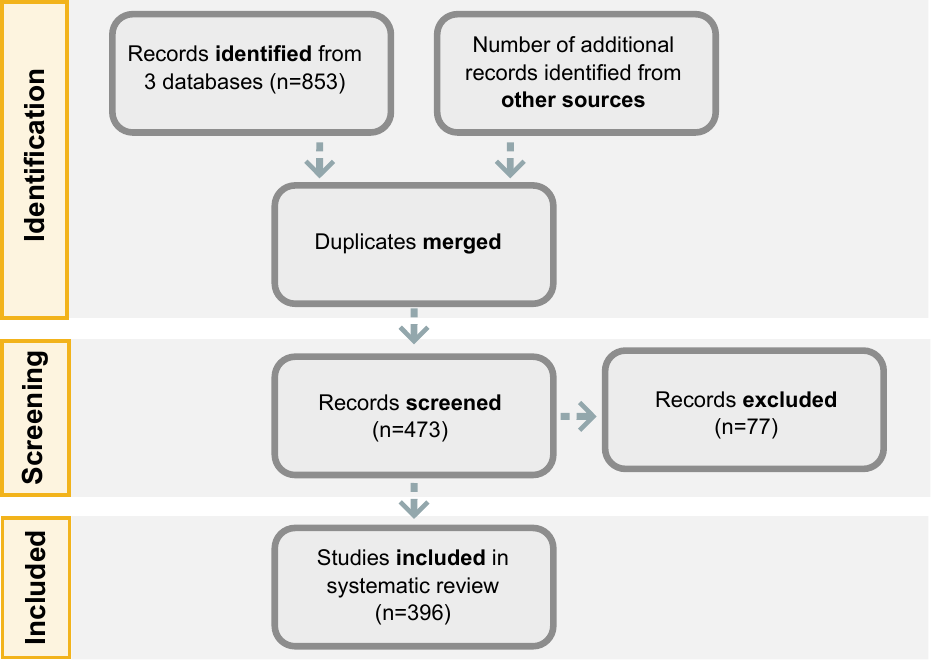}
    \caption{Diagram that outlines our literature review method according to PRISMA guidelines.}
    \label{fig:prisma}
\end{figure}

Abstracts of full-text articles were screened and sought for retrieval if their main subject were \acp{NCA}.
We follow our \ac{NCA} definition as presented in Sec.~\ref{sec:nca_definition}.
Finally, full-text articles were excluded that only mentioned \acp{NCA} as a side note while not actually contributing to the field.
The remaining papers were manually tagged with the respective downstream tasks, datasets, and domains.
Our workflow is illustrated in Figure~\ref{fig:prisma}.
We further used the tagged metadata records to analyze the frequency of different datasets and downstream tasks in which \acp{NCA} are applied.

\subsection{Limitations}
Studying applications and new architectural improvements for \acp{NCA} is a rapidly growing research field.
Following a general trend that can be observed especially in deep learning and \ac{AI}, 58 of the research papers are \textit{arXiv} preprints that have not (yet) undergone peer review.
However, some of the publications presented here are backed by public codebases, making the presented findings reproducible to at least some degree.
We therefore decided to include \textit{arXiv} papers in our review, but put an emphasis on findings discussed in peer-reviewed papers.

\subsection{NCA Definition}\label{sec:nca_definition}

We define an \ac{NCA} as an iterative cellular computation in which each cell maintains a state and updates that state using a local, parameterized update rule.
The update rule may depend on the state of the cell and on information obtained from a defined local neighborhood (the neighborhood must not be the entire image). 
At least part of this recurrent cell-update rule must be parameterized by continuous-valued weights that are not constrained to a discrete set.
The same update rule, or rules derived from a shared parameterization, is applied repeatedly across cells and time.

Cell states may be discrete or continuous and may contain additional persistent latent variables.
The update function need not contain an intermediate hidden representation. 
If transient hidden representations are used, these are local and must not be shared with other cells (otherwise the model becomes a recurrent \ac{CNN}).
Persistent latent variables through which cells communicate are considered part of the cell state.
We impose no restrictions on the topology of the cellular domain, which may be regular or irregular.

Furthermore, we require rollout-aware training.
Each parameter-update step must be based on a rollout containing at least 2 consecutive applications of the NCA update rule.
The state produced by one NCA step must serve as input to the subsequent NCA step, such that the optimization objective depends on the model's multi-step dynamics rather than exclusively on isolated one-step transitions.
This requirement is independent of the optimization algorithm.
For gradient-based training, the optimization signal must account for the dependency through at least 2 consecutive NCA steps, for example through \ac{BPTT}.
For gradient-free training, candidate parameterizations must analogously be evaluated according to an objective computed from a rollout of at least two consecutive NCA steps.

\section{Fundamentals}\label{sec:fundamentals}

The idea of \acp{CA} was first formalized as a discrete computation model by \citet{neumann1966theory}.
Later, John Conway created a practical manifestation of \acp{CA}, called Conway's Game of Life~\citep{gardner1970mathematical}.
Conway and his colleagues tried to investigate how a given initial formation would behave, to little success. 
It was later shown that Conway's Game of Life is in fact Turing complete, and that its halting problem is undecidable. 
Since then, \acp{CA} have been used for modelling differential equations, such as tumor growth dynamics~\citep{valentim2023cellularautomaton}.

\acp{NCA} are a new type of neural architecture that marries the ideas of two bio-inspired systems, by enhancing \ac{CA} with neural networks.
Typically, \acp{NCA} work on an image grid with an extended channel dimension, where a common, learned local rule is applied to each of the cells in an iterative fashion.

\begin{figure}
    \centering
    \includegraphics[width=0.8\linewidth]{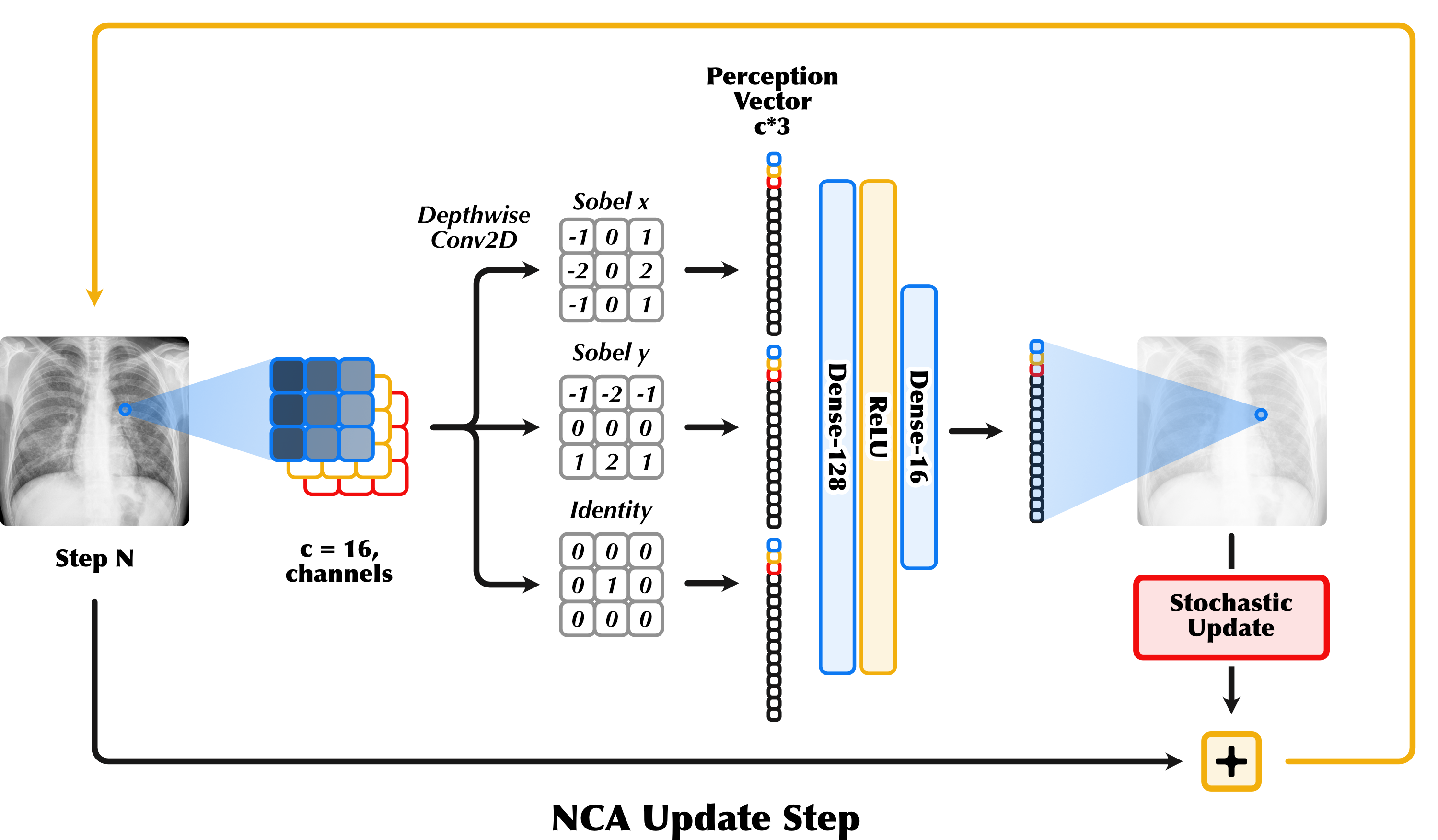}
    \caption{The original \ac{NCA} architecture (as proposed by~\citet{mordvintsev2020growinga}) consisting of predefined or learned convolutions for perception, $1\times1$ convolutions, and pixel-based dropout for the adaptation.
}
    \label{fig:nca}
\end{figure}

A good basis for understanding \acp{NCA} is the paper (and interactive online demo) by \citet{mordvintsev2020growinga}, who initially proposed the type of \acp{NCA} that this review targets.
Prior to their work, which was published in 2020, other algorithms described as \textit{"Neural Cellular Automata"} were proposed, but these were often grounded on vastly different concepts~\citep{boehme1992neural}.
In the following explanations, we will utilize the terminology and basic architecture described by \citet{mordvintsev2020growinga}, as it serves as a common ground for most recent works on \acp{NCA}.
Figure~\ref{fig:nca} illustrates the interplay of building blocks for our canonical \ac{NCA} definition.

An \ac{NCA} operates on a lattice of cells, iteratively applying a fixed local update rule as in a standard \ac{CA} framework~\citep{neumann1966theory}.
Each cell state is represented by a $C$-dimensional real vector.
We name the entirety of all cells the \textit{state}, represented by $\textbf{S}\in\R^{C\times H\times W}$ where $H$ and $W$ are the dimensions of the grid.
The \ac{NCA} encodes the update rule in a neural network, computing the next state $\textbf{S}^{t+1}_{C\times H\times W}$ of time $t+1$ given the previous state $\textbf{S}^{t}_{C\times H\times W}$.
The \ac{NCA} update function can be divided into different components, which we explain separately.

\noindent\textbf{Perception:} The perception mechanism is responsible for aggregating the states of its neighboring cells.
Specifically, neighborhood values are aggregated separately for each of the $C$ dimensions as a weighted sum of each neighbor.
Implemented as depthwise convolutions, the weights are either hard-coded as proposed by \citet{mordvintsev2020growinga} or learned as by \citet{kalkhof2023mednca}.
At least in the hard-coded perception, \textit{multiple} aggregation matrices (convolution kernels) are specified, such as Sobel filters in the x- and y-directions or Laplace filters.
Also, the identity of the current cell is preserved and concatenated with the results of the previous neighborhood aggregations.
For $n$ convolutional kernels, the result is a $(n+1)\cdot C$-dimensional perception vector for each cell in the grid $\textbf{Z}^t_{(n+1)C\times H\times W}$.
There is no general constraint on the number of aggregation matrices, so the perception vectors' dimensionality can be any multiple of $C$.
The perception mechanism is efficiently implemented by a depthwise convolution.

\noindent\textbf{Adaptation:}
The perception vectors are fed through a fully connected neural network.
Usually, the adaptation network consists of 2 linear layers with hidden dimensionality $D$ and a \ac{ReLU} activation function in between.
The last linear layer projects the latent representation down to the $C$-dimensional additive update.
The adaptation network can be efficiently implemented using $1\times 1$ convolutions.

\noindent\textbf{Stochastic Update:}
Cells do not update synchronously; instead, depending on the fire rate $\alpha\in(0,1]$, update vectors are randomly set to $0\in\R^C$ by probability $\alpha$.
The stochastic cell update can be interpreted as a cell-based dropout~\citep{mordvintsev2020growinga}.
The stochastic update is efficiently implemented by sampling an activation matrix $M\sim\textbf{Ber}(\alpha)$ from a Bernoulli distribution and multiplying it with the previous output.
Setting $\alpha=1$ results in a deterministic model with synchronous updates.

\noindent\textbf{Residual Connection:}
The computed update grid is added to the current state of the cell, introducing an additive residual connection.

\noindent\textbf{Seeding \acp{NCA}:}
After initializing the $C\times H\times W$ grid with height $H$, width $W$, and cell dimensionality $C$ (as above), the grid must be initialized with some information, which may depend on the task.
Hence, for imaging tasks, the state partitions into the image, the output, and hidden dimensions. 
In the task of growing images from a single seed, most of the grid can indeed remain empty, but a seed must be placed in the center of the grid to start the computation \citep{mordvintsev2020growinga}.
Placing multiple single-pixel seeds can be used to condition the NCA on different shapes, structures, and orientations of the generated pattern~\citep{randazzo2023growing}.
If the NCA should compute a result directly depending on the input, the grid is seeded with that specific input.
For example, in image segmentation, the input image is inserted in one of the channels of the initial grid.
The remaining channels of the initial grid can be initialized with zeros or uniform noise~\citep{pajouheshgar2024noisenca}.

\subsection{Architectural Variations}\label{sec:architecture_variations}
\noindent Several variations of the original \ac{NCA} recipe were proposed in follow-up work to \citet{mordvintsev2020growinga}, some of which are outlined in the following.

\noindent\textbf{ViTCA:}
Potentially, the biggest and most prominent change to the standard \ac{NCA} is the Vision Transformer Cellular Automata (ViTCA), which leverages the idea of cellular automata but replaces the fully connected backbone and neighborhood aggregation with a local transformer operating on the same local Moore neighborhood~\citep{tesfaldet2022attentionbased}.
The model was extended to a diffusion transformer~\citep{elbatel2024organisma}.

\noindent\textbf{Immutable state channels:}
Most \acp{NCA} applied to discriminative tasks initialize the state with the input and keep those state channels immutable.
The first work to incorporate this change is the segmentation \ac{NCA} by \citet{sandler2020image}.

\noindent\textbf{Normalization layers:}
Some \ac{NCA} architectures employ normalization layers, such as a global batch norm~\citep{kalkhof2023m3dnca} employed in the latent space of the \ac{NCA} feed-forward network (i.e. between the hidden layers).
\citet{kalkhof2024frequencytime} employ a cell-wise normalization to adhere to the cell-wise updates of \acp{NCA}. \citet{tesfaldet2022attentionbased} use layer normalization. Other normalization schemes can be used as well.

\noindent\textbf{Multiplicative Conditioning:}
\citet{sudhakaran2022goalguided} employ an embedding that is multiplied to the state channels in the \ac{NCA} grid.
Similarly, \citet{kalkhof2024frequencytime} and \citet{ranem2025ncadapta} multiply the embedding to the perception vector and the latent space of the \ac{NCA} adaptation network.

\noindent\textbf{$\mu$NCA:}
\citet{mordvintsev2021mnca} propose an even smaller \ac{NCA} architecture consisting of only convolutional kernels, a learnable non-linearity, and a single linear layer projecting the perception vector down to the dimensionality of the cells.
These measures allow shrinking the \ac{NCA} model to as few as 68 parameters.

\noindent\textbf{Positional encoding:}
\citet{pajouheshgar2024noisenca, kalkhof2025parameterefficient} include a global positional encoding that is appended to the perception vector in each \ac{NCA} update step. Similarly, ViTCA employs a positional encoding~\citep{tesfaldet2022attentionbased,elbatel2024organisma}

\noindent\textbf{Isotropic / Chiral \ac{NCA}:}
\citet{mordvintsev2022growing} proposed IsoNCA, an isotropic (rotationally- and mirror-equivariant) \ac{NCA} by replacing the Sobel filters with a Laplacian filter that satisfies this property. 
Later, \citet{randazzo2023growing} propose the steerable \ac{NCA}, which is a chiral (rotationally equivariant, but not mirror-equivariant) \ac{NCA} architecture. 
The steerable \ac{NCA} automatically infers the rotation of the Sobel filters at each step. 
Due to the stochastic nature of \acp{NCA}, the same \ac{NCA} may generate different rotations of the same image it was trained on.

\noindent\textbf{Non-regular input lattice:}
While the \acp{NCA} mentioned before operate on a regular 2D or 3D lattice with 8 or 26 neighbors, respectively, \acp{NCA} can be extended to irregular neighborhoods with a varying number of neighbors as a Graph NCA~\citep{grattarola2021learning}.
Similar to the message passing scheme~\citep{you2020design}, messages from its neighbors are summed and processed to update each cell's latent state. 
\citet{gala2023enequivariant} extend upon this early work, including a hidden state for each cell.
Finally, \citet{pajouheshgar2024mesh} propose MeshNCA operating on a 2-dimensional manifold mesh for texture generation.
They propose a mesh perception module based on spherical harmonics that extends Laplacian and Sobel filters to non-regular grids.

\noindent\textbf{ANCA:}
Most \ac{NCA}-architectures employ a static cell-lattice, where the spatial position of individual \ac{NCA}-cells is fixed in relation to the input data. \citet{kvalsund2026sensora} propose Active \ac{NCA} (ANCA), an \ac{NCA}-based image classification architecture that decouples the input-space from the \ac{NCA} state-space. The authors employ a cell grid that is spatially detached from the input data. Through dedicated channels in the state-space, individual cells can freely control the position of their receptive field in the image-space throughout the inference process.

\subsection{Loss Function and Training of NCAs}
\acp{NCA} can be trained using the same loss functions as other neural networks.
Depending on the task (segmentation, classification, regression), common loss functions are (binary) cross-entropy or mean squared/absolute error.
However, cell states must be aggregated for loss computation in tasks that yield a lower-dimensional prediction (e.g. image classification), for instance through average or maximum pooling followed by a fully connected head layer~\citep{deutges2024neurala}, or a majority vote of all cells~\citep{dohler2008cellular}.

Due to the recurrence, \acp{NCA} are trained by \acp{BPTT}, which unrolls the computational graph during the forward pass (over time) and accumulates the gradients obtained in each time step.
The unrolled computational graph can become long for a large number of NCA iterations, which renders training slow and puts high demand on the GPU memory.

Other methods train \acp{NCA} with gradient-free optimization algorithms, e.g., genetic algorithms~\citep{dohler2008cellular}.
However, these are exceptions. The preferred way of training \acp{NCA} is via gradient-based optimization.

\subsection{Relationship to Common Neural Models}
\begin{figure}
    \centering
    \includegraphics[width=0.9\linewidth]{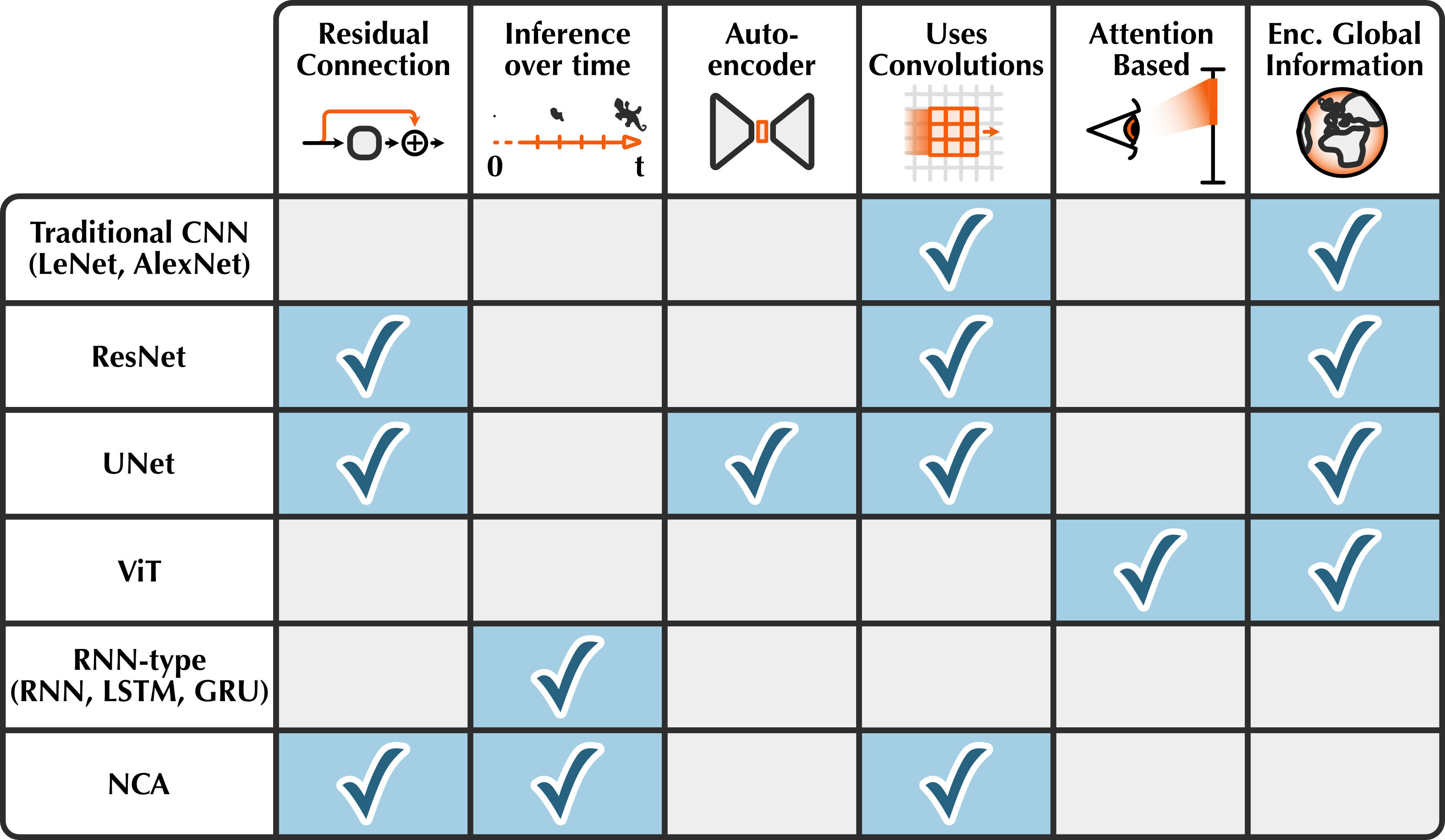}
    \caption{Relationship of common neural architectures to \ac{NCA} models. $^1$:~CNN-type models may operate locally if they are shallow.}
    \label{fig:modelComparison}
\end{figure}

Although \acp{NCA} make use of convolutions and neural networks, the approach is still rather different from that of \acp{CNN}. 
Typically, \ac{CNN}-based architectures like ResNet~\citep{he2016deep} are much larger than \ac{NCA} models, and they have been shown to have high parameter redundancy~\citep{denil2013predicting}, whereas \ac{NCA} weights are usually more compact.
While \acp{CNN} operate in a single forward pass, \acp{NCA} apply multiple forward passes utilizing the same set of filters.
\citet{mordvintsev2020growinga} describe \acp{NCA} as \textit{"Recurrent Residual Convolutional Networks with ‘per-pixel’ Dropout"}.

\acp{ViT} and ViTCA models exhibit the same differences that exist between \acp{CNN} and \acp{NCA}.
The ViTCA is iterated multiple times, utilizing the same set of parameters.
Similar to the description for NCA, ViTCA can be seen as a recurrent convolutional transformer with 'per-pixel' Dropout.

Figure~\ref{fig:modelComparison} shows the similarities and differences of \acp{NCA} with other established deep learning architectures.

\subsection{Composite NCA architectures}
While a single \ac{NCA} already proves capable of solving different tasks, some methods combine multiple \acp{NCA}, potentially with other architectures, to create an even more expressive model.
Multiple \acp{NCA} can be applied sequentially at increasing resolutions to create a model scalable to high-resolution training data~\citep{kalkhof2023mednca,kalkhof2023m3dnca}.
\citet{yue2024skina} and \citet{menta2024latenta} embed an \ac{NCA} in a UNet-style autoencoder architecture.
The \ac{NCA} performs the computation at the autoencoder's bottleneck.
The remaining layers are responsible for down- and upscaling the input and output of the \ac{NCA}.
\citet{xu2024adanca} include an \ac{NCA} in a vision transformer operating directly on a grid of tokens. The \ac{NCA} improves the model's robustness to noisy images and adversarial attacks.

\subsection{Related concepts and other names of NCA}
There are several related concepts to \acp{NCA}.
Some concepts share similarities with \acp{NCA}, even fitting our definition in Section~\ref{sec:nca_definition}, but utilizing a different name, while others do not fit our definition, despite being very close.
Here, we review those concepts.
Tab.~\ref{tab:related_concepts} lists those methods and their main differences from \acp{NCA}.

\ac{CA} and \textit{probabilistic \ac{CA}} use hand-engineered rules, while the latter also implements a probability distribution over state transitions.
The probabilistic \ac{CA} is also referred to as \ac{Markov-CA}, \mbox{Markov Chain CA}, \mbox{CA-Markov}, and \mbox{CA-MC}.

While \ac{CLA}~\citep{beigy2004mathematical} sounds related to \acp{NCA}, their concepts are rather different.
In \ac{CLA}, each cell has its own set of rules.
So, instead of learning a shared rule, each cell optimizes its own rule.

The idea of \ac{ANN-CA}~\citep{li2002neuralnetworkbaseda} is somewhat similar to that of \acp{NCA}, as it combines a neural network with a \ac{CA}. 
However, the neural network is trained only for a single \ac{CA} transition.
Only after training is the \ac{ANN} embedded into the \ac{CA} to enable longer rollouts.
Subsequently, \ac{ANN-CA} does not have hidden states, and the output is always a discrete variable (not warranted by our definition). 
The \ac{ANN-CA} exhibits high popularity in \ac{LULC} modeling~\citep{zhang2024predictiona,daich2025spatiotemporala,otmanecherif2026nonlinear}.

\ac{ANN-CA-MC} \citep{gharaibeh2020improvinga} combines the \ac{ANN-CA} with a Markov chain stage.
The Markov chain estimates pairwise class transition probabilities and converts them into a target number of cells to convert per class pair at each \ac{CA} step.
The \ac{ANN-CA} component independently produces a per-cell transition-potential score.
Subsequently, the scores are ranked by this score and converted until the Markov-derived quota is met.

The \ac{LSTM-CA} \citep{liu2021simulation,rezaie2026robust} is another model from the \ac{LULC} direction. 
It integrates an \ac{LSTM} module into the \ac{ANN-CA}. 
An LSTM model is trained on the temporal evolution of its state and those of its neighbors.
After learning, the \ac{LSTM} is deployed in a \ac{CA}. 
No cell-to-cell communication is learned.
Each cell aggregates only the input variables from its neighbors.

\ac{ACAN}~\citep{matsubara2013asynchronous} is a single neuron that implements a transition function with learned parameters.
The \ac{ACAN} is optimized via a gradient-free hill-climbing algorithm over the entire trajectory.
However, as it is a single neuron, the neuron is not embedded in a topology but rather works in isolation.
The \ac{ACAN} is more similar to a recurrent neuron.

\ac{ANCAN} \citep{matsubara2016asynchronous} implements multiple \acp{ACAN} in fixed a topology to build a neural network.
Rather than encapsulating the input (e.g., an image, connectivity graph), the topology represents connectivity in the neural network structure.
While this architecture works substantially different to \acp{NCA} by generating the prediction over space (from input neurons to output neurons), rather than over time, this architecture does fit out \ac{NCA} definition.
Since \ac{ACAN} neurons change values over time, the \ac{ANCAN} predictions also emerge over time, rather than after a fixed time, as in a conventional neural network.  

Cellular Neural Network (CNN, not to be confused with \textit{Convolutional Neural Network}) \citep{chua1988cellular,dohler2008cellular} matches our \ac{NCA} definition cleanly. 
It consists of 2 state channels (1 input and 1 output).
The learned update rule is parameterized by 19 parameters. 
The CNN is optimized with a genetic algorithm over 50 rollout steps.
The final classification score is obtained by majority vote of all cells.

\citet{omija2026architectural} already realized the structural similarity between \ac{ConvLSTM}~\citep{shi2015convolutional} and \ac{NCA}.
\ac{ConvLSTM} is a convolutional \ac{LSTM}.
Instead of a simple \ac{MLP}, the update network uses an advanced \ac{LSTM} architecture with specific memory management gates.
The hidden state representation of the \ac{LSTM} is shared with its local neighborhood.

\begin{table}[t]
    \centering
\begin{tabular}{l|l|l}
\toprule
    \multirow{3}{*}{\rotatebox{90}{NCA?}}\\
&\\
     & Framework & Difference to \ac{NCA}\\ 
     \midrule
     \multirow{7}{*}{\rotatebox{90}{No}}
     &\ac{CA}& Rules are hand-engineered\\
     &Prob. \ac{CA} & Rules are hand-engineered\\
     &\ac{CLA} & Each cell has its own rule, no rollout-aware training\\
     &\ac{ANN-CA} & Only trained for one step\\
     &\ac{ANN-CA-MC} & Only trained for one step\\
     &\ac{ACAN} & Single neuron, no topology\\
     &\ac{LSTM-CA} & No learned cell-to-cell communication\\
     
     \midrule
     \multirow{3}{*}{\rotatebox{90}{Yes}}
     &\ac{ANCAN} & Prediction emerges over space \textit{and} time\\
     &CNN & Gradient-free optimization\\
     &\ac{ConvLSTM}& Advanced memory and update mechanisms\\
     \bottomrule
\end{tabular}
    \caption{Other related concepts and their major differences from \ac{NCA}. Concepts are classified into whether they are an \ac{NCA} or not as warranted by our definition (Sec.~\ref{sec:nca_definition}).}
    \label{tab:related_concepts}
\end{table}

\subsection{NCAs on small-scale and specialized hardware}

Thanks to their decreased number of parameters and good balance between model size and accuracy, NCAs have gained attention by researchers investigating deployment on embedded platforms.
Successful deployment of \acp{NCA} was reported on single-board computers (Raspberry Pi, smartphones), and even on microcontroller hardware (\mbox{ESP32-S3}).
Further, experimental approaches for creating self-organizing cellular devices exist.

In most of these scenarios, no additional pruning or dedicated techniques for model size reduction had to be employed, delivering similar accuracy to a PC-scale deployment.
However, the maximum input image size is a bottleneck for deployment on embedded platforms which are often memory-constrained.
Table~\ref{tab:comparisonHardware} shows a comparison of different \ac{NCA} implementations on various embedded hardware platforms, which will be detailed in the following:

\begin{landscape}
\begin{table*}[t]
    \centering
    \begin{tabular}{l l l l c}
    \toprule
        Paper & Platform & Model & Application & Train \\
        \midrule
        \citet{walker2022physicala} &  Arduino & Simplified NCA & Self-organizing digit classification & No \\
        \citet{kalkhof2023mednca} & Raspberry Pi B+ &  Multi-scale NCA & MR segmentation (2D) & No \\
        \citet{kalkhof2023m3dnca} & Raspberry Pi 4 B (2GB) & Multi-scale 3D NCA & MR segmentation (3D) & No \\
        \citet{kalkhof2024unsupervised} & Smartphones & Med-NCA & X-ray segmentation & Yes \\
        \citet{li2024deepa} &  Photonic hardware & Photonic NCA & FashionMNIST classification & Possible \\
        \citet{krumb2025encapsulatea} & ESP32-S3 & NCA & Endoscopic segmentation, depth estimation & No\\
        \citet{dudziak2021neural} & Focal Plane Sensor Processor & Quantized NCA & Self-organizing digit classification & No \\
        \citet{lemke2025equitablea} & Smartphones & Med-NCA & X-ray/Ultrasound segmentation & Yes \\
        \citet{lemke2025octreenca} & Raspberry Pi 4 B (2GB) & Multi-scale 3D NCA & X-ray/RGB/Microscopy segmentation & No
    \end{tabular}
    \caption{Works that present NCA on different embedded hardware platforms. "Train" indicates whether training is also carried out on the small-scale platform, or only the inference code is ported.}
    \label{tab:comparisonHardware}
\end{table*}
\end{landscape}

\noindent\textbf{Single-board computers:}
\acp{NCA} were successfully deployed for inference on single-board computers such as the Raspberry Pi~\citet{kalkhof2023mednca,kalkhof2023m3dnca,lemke2025octreenca}.
\citet{kalkhof2023mednca} demonstrated an \ac{NCA} for $320\times 320$ pixels slice-wise \ac{MRI} segmentation 
on a Raspberry Pi (1st generation) B+ with a 1~GHz single-core SoC and 512~MB RAM.
In their experiment, inference takes roughly 30 minutes, compared to few seconds on consumer-grade GPU hardware.

\noindent\textbf{Smartphones:}
Smartphones are ubiquitous hardware, and they are even available in resource-constrained environments.
This makes them particularly interesting for point-of-care image analysis~\citet{kalkhof2024unsupervised} or larger-scale federated learning~\citet{lemke2025equitablea}, enabled by \acp{NCA}.

\citet{nunn2026efficient}~report an inference speed of \SI{3.7}{\milli\second} on a Moto G Stylus 5G 2023 smartphone.

Whereas most approaches concern with inference on mobile devices, \citet{kalkhof2024unsupervised} \textit{train} \acp{NCA} on smartphones for 2D X-ray segmentation.
The implementation leveraged TensorFlow Lite for mobile deployment.
Depending on the smartphone model, training may take between 38 to 82 hours total (\SI{91.2}{\second} to \SI{196.8}{\second} per epoch at 1500 epochs).
However, the presented approach allows for efficient unsupervised site-specific finetuning, which roughly takes 5 to 12 hours total.
The study focused particularly on affordable and refurbished Android smartphones.

Federated learning approaches manage to relax the computational requirements on smartphones through decentralization, allowing to train \acp{NCA} on edge devices in a distributed topology~\citet{lemke2025equitablea}.
The proposed setup takes less than \SI{2.5}{\second} per epoch on a single node.

\noindent\textbf{Microcontrollers:}
\citet{krumb2025encapsulatea} propose \acp{NCA} for capsule endoscopic segmentation and depth estimation, which are ported to the ESP32-S3 microcontroller.
An optimized inference loop and \ac{SIMD} instructions reduce inference time from \SI{9}{\second} to \SI{3}{\second} per image.
The implementation makes use of the ESP32-S3's floating point unit, meaning that no quantization needs to be applied for deployment on microcontroller hardware.

\citet{walker2022physicala} propose an \ac{NCA} based on physical hardware tiles, each of which is a \ac{PCB} with an Arduino Mega 2560.
These hardware tiles communicate to their adjacent tiles via UART in order to aggregate their physical neighborhood for self-classification.
The authors deploy an \ac{NCA} rule for digit classification and physically align the tiles in digit shapes which are then classified.
The authors suggest that this idea of shape-awareness could be useful for automatic damage detection of complex structures.

\noindent\textbf{Photonic Circuits:}
\citet{li2024deepa} propose a \ac{PNN} implementation of an \ac{NCA}.
To our knowledge, this is the first (and so far only) implementation of \acp{NCA} in analog hardware.
The proposed in-silico setup is capable of training and inference, and is validated on FashionMNIST ($99.3\%$ reported test set accuracy).

\noindent\textbf{Efficient GPU implementations:}
General-purpose deep learning frameworks like PyTorch or tensorflow do not necessarily yield the best-performing \ac{NCA} implementation, especially since these frameworks are optimized for layer-wise inference.
\citet{lemke2025octreenca} propose a custom CUDA kernel for OctreeNCA, which performs cell-wise (rather than layer-wise) inference, therefore reducing \ac{VRAM} requirements significantly.

Another approach, $\mu$NCA, was described earlier in this paper (Section~\ref{sec:architecture_variations}), \citet{mordvintsev2021mnca}.
$\mu NCA$ reduces overall \ac{NCA} complexity by simplifying the nonlinearity and thus reducing parameter count, allowing for a concise and efficient implementation as a fragment shader (demo: \url{https://www.shadertoy.com/view/7cXSDX}).

\noindent\textbf{NCA as Digital Circuits:}
\citet{miotti2025differentiable} propose DiffLogic \ac{CA}, which are \acp{NCA} leveraging \acp{DLGN} as their update rule.
The authors identify the potential for utilizing \ac{DLGN} in order to create an efficient implementation of \ac{NCA} for \acp{FPGA} or \acp{ASIC}.

However, the logic gate configuration can also be modelled as a self-organizing system akin to \acp{NCA}.
\cite{bena2026selforganising} propose a setup with logic gates that change states based on neighborhood configurations, yielding self-assembling and self-healing digital circuits.

\noindent\textbf{Self-recognizing Cellular Devices:}
A fundamental idea for the development of \ac{CA} was the \textit{universal constructor} designed by John von Neumann in the 1940s~\citet{neumann1966theory}, as a prototype model for self-assembling, self-reproducing or self-repairing devices.
This idea reflects in recent work on self-recognizing cellular physical devices based on microcontroller hardware.
In-silico simulators and visualizations are proposed, such as \citet{woiwode2025rotationinvariant}.
A physical approach was demonstrated in hardware for simple 2D shapes (digits) by \citet{walker2022physicala}.
\citet{moreno2026smarta} propose a hardware design for 3D self-recognizing hardware bricks, and conduct extensive simulation studies demonstrating abilities to self-classify and self-heal complex 3D structures.
Related studies investigate the potentials of \acp{NCA} for self-repairing soft robots in computer simulations \citet{horibe2021regenerating,horibe2022severe}.

\section{Applications of NCAs}\label{sec:applications}

\begin{landscape}
\begin{table*}
    \centering
    \begin{tabular}{lllll}
    \toprule
        Paper &  Task & Modality & Anatomy & Key idea \\
         \midrule
         \citet{kalkhof2023mednca} &  Segmentation & MRI & Prostate, Hippocampus & 2-level architecture\\
         \citet{kalkhof2023m3dnca} &  Segmentation & MRI & Prostate, Hippocampus & Multilevel 3D architecture\\
         \citet{kalkhof2024unsupervised}&Segmentation& X-Ray&  Lungs &Unsupervised Fine-tuning\\
         \citet{ali2024breasta}&Segmentation&Mammography&Breast Tumor&M3D-NCA + Shape guidance\\
         \citet{korevaar2024generalization}& Segmentation & OCT, MRI & Retina, Cardiac& Investigate robustness to shift\\
         \citet{yue2024skina}& Segmentation& RGB &Skin lesions& latent NCA on shallow UNet\\
         \citet{mittal2025medsegdiffncaa}&Segmentation& RGB &  Skin lesions & Diffusion-based Segmentation\\
         \citet{ranem2025ncadapta}&Segmentation & MRI & Hippocampus&Continual Learning\\
         \citet{kalkhof2025mednca}& Segmentation & MRI, CT, U/S,  & Hippocampus, Prostate, &Visualization of hidden states\\
         &  &  X-Ray, RGB& Liver, Breast, Skin, &\\
         &  &  & Heart, Lung&\\
         \citet{lemke2025equitablea}&Segmentation& Ultrasound, X-Ray& Spleen, Lungs, Fetal liver & Federated Learning\\
         \citet{lemke2025octreenca}&Segmentation&MRI, RGB, Microsc. &Prostate, Surgery, Path. & Multiscale NCA+CUDA layer\\
         \citet{silbernagel2025rnca}& Seg. Refinement& CT, OCT, MRI & Liver, Retina, Heart & NCA repairs segm. errors\\ \citet{deutges2025neural} & Segmentation & Microscopy & Blood Cells & Train from class label only\\
         \citet{sadafi2026measuring} & Segmentation & RGB & Endoscopic & Confidence/uncertainty est. \\
         \citet{murgu2026pderefined} & Segmentation & RGB & Retina & Swin-UNet-Backbone+NCA head \\
         
         \midrule
         \citet{dohler2008cellular} & Classification & MRI & Hippocampus & Gradient-free optimization \\
         \citet{deutges2024neurala} & Classification & Microscopy &Blood Cells & NCA+Max-Pool classifier \\
         \citet{yang2025attention} & Classification & Microscopy & Blood Cells, Colon, & NCA+attention pooling\\
         && & Lymph Node, Urine & \\
         \citet{yang2026hierarchical}& Classification & Microscopy &Blood Cells, Urine & 2-level classification NCA\\
         \midrule
         
         \citet{ranem2024ncamorph} & Registration& MRI & Hippocampus, & NCA + VoxelMorph\\
         && & Prostate, Brain & \\
         
         \midrule
         
         \citet{krumb2025encapsulatea}& Depth Estimation & RGB & Capsule endoscopy & Depth estimation with NCA \\
         
         \midrule

         \citet{lemke2026sterilizable}&Scene Graph Gen.&RGB&Surgery & NCA + relation predictor\\

         \midrule
         
         \citet{manzanera2021patientspecific}& Nodule Growth Sim. & CT& Lung & NCA grows nodules \\
         \citet{elbatel2024organisma}& Diffusion & Fundus, OCT & Retina & Diff. Transformer+NCA \\
         \citet{kalkhof2024frequencytime}& Diffusion & Microscopy & Pathology & Diff. NCA in Fourier space\\
         \citet{lang2025temporal}& Post-contrast gen. & MRI & Breast & Time-continuous generation  \\
         \citet{schwarz2026mammographic}& Diffusion&X-Ray&Breast&GeCA on Mamography\\
         \citet{luu2026coarse}& GAN & Microscopy & Colon, Blood Cells & NCA + StyleGAN\\
    \end{tabular}
    \caption{Overview of works utilizing NCAs for medical image analysis}
    \label{tab:ncaformed}
\end{table*}
\end{landscape}

In this section, we review popular application areas and the value provided by the \acp{NCA} in contrast to other model architectures.
We review \acp{NCA} for several medical image analysis tasks, robotics, simulation, and image and texture generation.

\subsection{Medical Image Analysis}

In this section, we present various works that utilize \acp{NCA} for different medical image analysis tasks.
Medical image analysis is dominated by large convolution- or transformer-based architectures achieving impressive quality across a wide range of applications and anatomies.
These methods already pose great value for the healthcare sector.
However, the ability to solve tasks accurately is not the only concern in medical imaging.
Specifically, proper resource efficiency, robustness, and out-of-distribution handling are at least equally important for medical applications.

As the expressiveness of \acp{NCA} emerges only from local interactions, the model is inherently invariant to image properties such as size and scaling factor~\citep{kalkhof2023m3dnca}.
Also, \acp{NCA} are robust to imaging artifacts~\cite{kalkhof2023mednca}, while the \ac{NQM} provides an easy way to detect critical model failure~\citep{kalkhof2023m3dnca}.
The hidden channels of the \ac{NCA} provide an easy way for explainability~\citep{deutges2024neurala}.
Last, the minimal architecture of \acp{NCA} is parameter-efficient and lightweight, enabling training and inference on minimal hardware such as Raspberry Pis~\citep{kalkhof2023m3dnca} and microcontrollers~\citep{krumb2025encapsulatea}.

All these properties of \acp{NCA} provide value for medical image analysis beyond great task-solving capabilities.
To this date, \acp{NCA} have been explored on a variety of tasks and evaluated on different anatomies.
Table~\ref{tab:ncaformed} provides an overview of various works and respective downstream tasks in medical imaging.

\subsubsection{Medical Image Segmentation}

Medical image segmentation describes the task of delineating anatomical or pathological regions in a medical image.
In most of the recent works, medical images are segmented per-pixel or per-voxel in a binary or multi-label setting.
Such segmentations can be used for visualization purposes, to determine the size of pathologies, or for creating digital twins of patients for surgical planning~\citep{krumb2026patronus}.

The nnUnet framework~\citep{isensee2021nnunet}, which leverages the UNet architecture at its core~\citep{ronneberger2015unet}, can be deemed the current gold standard for data-driven medical image segmentation.
Typically, without employing techniques like pruning or quantization, the segmentation models trained by nnUnet reach sizes of several hundred Megabytes.

\begin{figure}[htbp]
    \centering
    \includegraphics[width=0.9\linewidth]{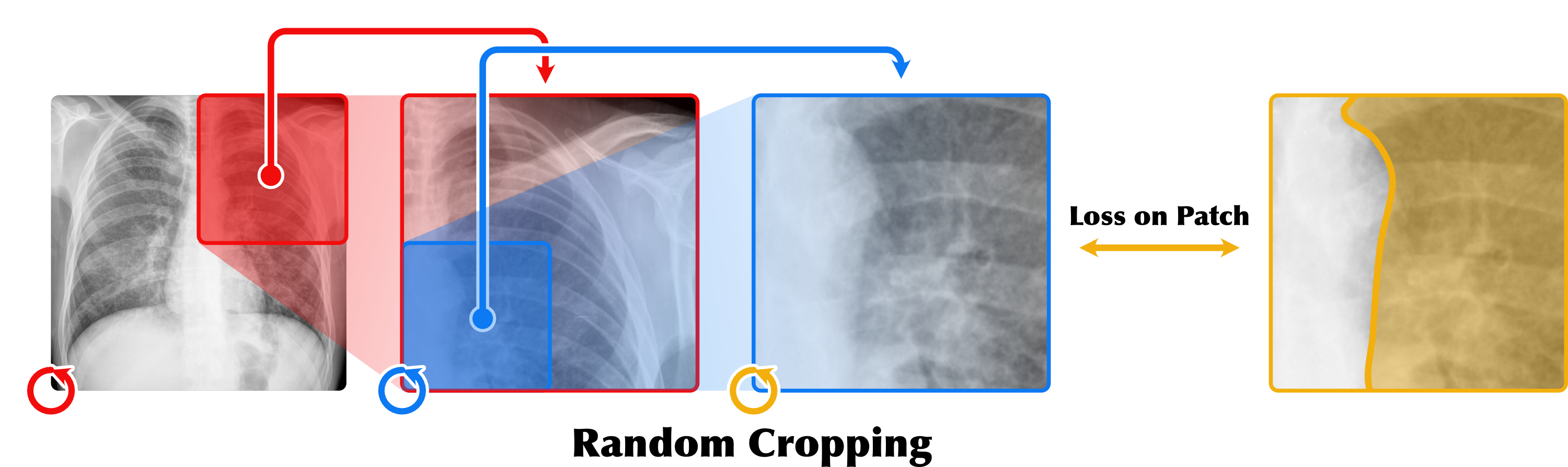}
    \caption{By randomly cropping on the high-resolution layers during training, MED-NCA enables training NCAs on large image sizes.}
    \label{fig:random_cropping}
\end{figure}

The first image segmentation algorithm based on \acp{NCA} was developed by \citet{sandler2020image}.
They use an \ac{NCA} to segment the foreground object in animal photos of resolution $96\times96$. As mentioned in Sec.~\ref{sec:architecture_variations}, they proposed leaving the input image immutable in the \ac{NCA} state. They also developed a reset component, which was not adopted in later works. While this method laid the foundation for discriminative modeling with \acp{NCA}, training was inherently difficult, as it required many steps to transfer global knowledge, resulting in memory requirements that exceeded the limits of recent \acp{GPU}.
This made the entire method unscalable for more complex tasks and high-resolution data. 
\mbox{Med-NCA}~\citep{kalkhof2023mednca} solves those issues by proposing a two-stage \ac{NCA} architecture.
A first \ac{NCA} transfers global knowledge at $1/4$ of the original spatial resolution.
The states are upscaled and concatenated with the high-resolution image.
The second \ac{NCA} predicts the final segmentation at the full resolution. During training, the high-resolution NCA is trained on random patches to further reduce the VRAM requirements (see Fig. \ref{fig:random_cropping}). Afterwards, it is applied to the full-scale image. 

\begin{figure}[htbp]
    \centering
    \includegraphics[width=0.9\linewidth]{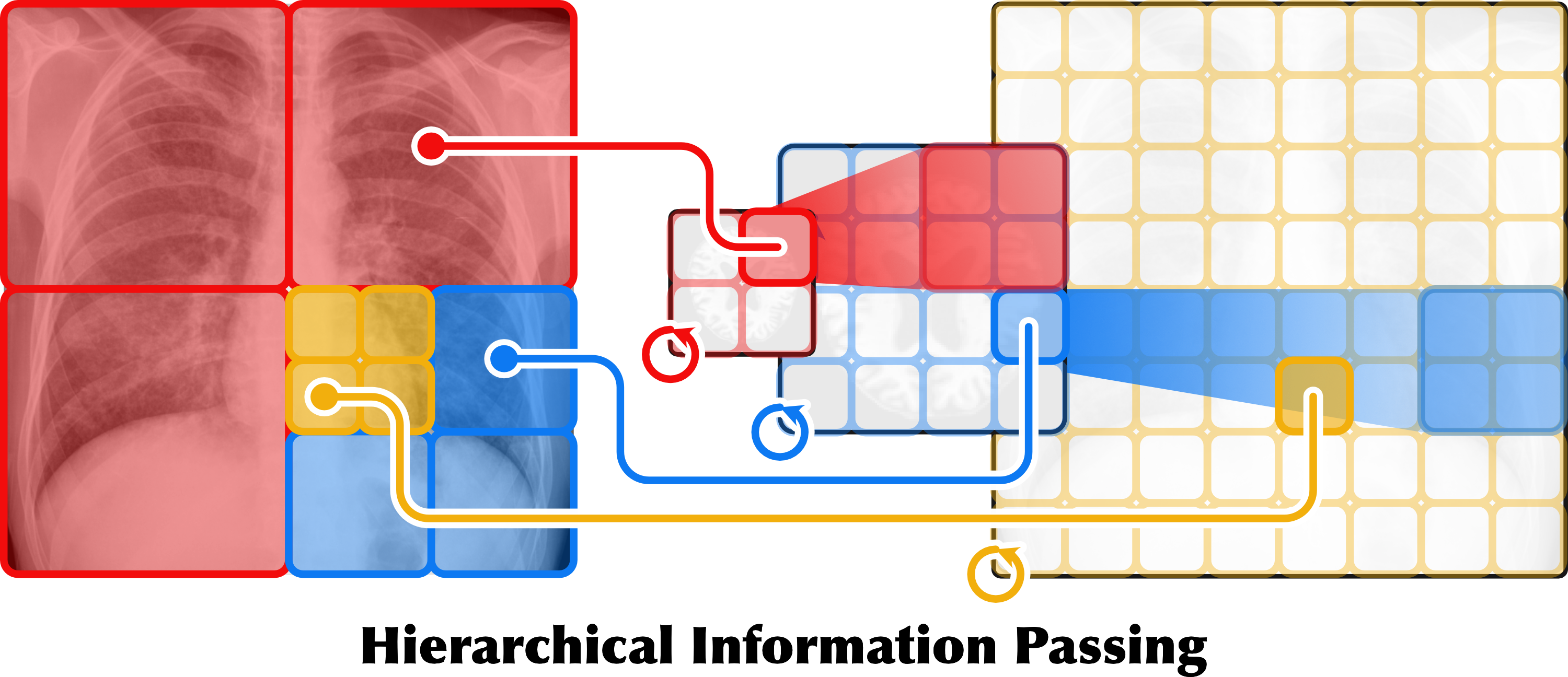}
    \caption{Passing information over multiple scales accelerates information flow in NCAs.}
    \label{fig:hierarchical_inf_pass}
\end{figure}

The architecture was further improved to utilize a variable number of \acp{NCA} and adapted to 3D segmentation in \mbox{M3D-NCA}~\citep{kalkhof2023m3dnca} (illustrated in Fig. \ref{fig:hierarchical_inf_pass}).

To stabilize the training of the non-deterministic \ac{NCA}, they utilize batch duplication.
Following the success of \mbox{M3D-NCA}~\citep{kalkhof2023m3dnca}, interest in \ac{NCA} for medical image segmentation increased, with more sophisticated or specialized works following.

\citet{yue2024skina} propose a different approach to solving the VRAM and convergence issues of NCAs when applied to high-resolution data. Specifically, they combine NCA with a shallow UNet. While the UNet performs up- and down-scaling of the input/output, the NCA is embedded in the UNet's latent space, performing the actual segmentation.

MedSegDiffNCA~\citep{mittal2025medsegdiffncaa} is a discriminative generative model that performs medical image segmentation using a diffusion model~\citep{wu2024medsegdiffv2}.

\citet{lemke2025octreenca} extend the M3D-NCA~\citep{kalkhof2023m3dnca}, leveraging $5$ NCAs applied in a sequence, each on a different scale. This scheme further reduces VRAM demands, increases runtime efficiency, and slightly improves segmentation quality.

\citet{lemke2025equitablea} propose FedNCA, a federated learning scheme based on Med-NCA~\cite{kalkhof2023mednca}. They highlight how FedNCA achieves similar segmentation accuracy to UNets~\citep{ronneberger2015unet} and TransUNets~\cite{chentransunet}, but with significantly lower communication costs.

rNCA is an \ac{NCA} for repairing segmentations. The model is trained to reverse the corruption of segmentation masks. At inference, the trained model repairs imperfect segmentations~\citep{silbernagel2025rnca}.

Last, \citet{deutges2025neural} train an NCA segmentation model from classification labels only. By utilizing the explainability of \acp{NCA}~\citep{deutges2024neurala}, the segmentation mask forms in the hidden states of the \acp{NCA}.

\subsubsection{Medical Image Registration}

\textit{Registration} describes the task of aligning multiple images of different modalities or subjects (patients) into a common coordinate system.
Traditionally, this task is approached by optimizing a linear transform that minimizes an error metric between the fixed and moving image.
Further, keypoint-based approaches allow for the estimation of deformation fields for elastic registration.
However, these approaches are costly and are thus replaced by learning methods.
With Voxelmorph~\citep{balakrishnan2019voxelmorpha} and follow-up work such as Hypermorph~\citep{hoopes2021hypermorph}, UNet-based methods were proposed that estimate an elastic deformation field quickly.
Further, TransMorph~\citep{chen2022transmorph}, ViTVNet~\citep{chen2021vitvnet}, and NICE-Trans~\citep{meng2023noniterative} were proposed as transformer-based architectures for elastic registration.

To the best of our knowledge, the only NCA-based registration approach was proposed by \citet{ranem2024ncamorph}.
The authors use a pipeline that is inspired by VoxelMorph~\citep{balakrishnan2019voxelmorpha}, using an \ac{NCA} for generating a deformation field.
Akin to VoxelMorph, the \ac{NCA} is trained with a smoothness penalty and a similarity loss.

\subsubsection{Endoscopic Depth Estimation}
The first work for estimating depth using \acp{NCA} was proposed by~\citet{krumb2025encapsulatea}.
The \ac{NCA} takes a single image from a capsule endoscope and predicts its depth map.
The model itself is small enough to be deployed on an ESP32-S3.
The model is trained using data pseudo-labeled from the foundation depth estimation model Depth Anything V2~\citep{yang2024depth}.

\subsection{Generation and Synthesis}
Generative models, such as \acp{GAN}, \acp{VAE}, or diffusion models, are well-established methods for image generation.
Beyond content generation, these models find various potential applications in visual computing, including image reconstruction, artifact removal~\citep{fuchs2024harp}, denoising~\citep{nazir2024recent}, or (un-)conditional synthesis~\citep{kazeminia2020gans}.
The latter can be utilized to generate additional out-of-distribution samples for training other models~\citep{du2021vos}, reduce label imbalance~\citep{frid-adar2018synthetic}, or to craft new unseen cases for training medical personnel~\citep {sivakumar2025sg2vid,frisch2025surgrid}.

While such generative models are already well-investigated, there are only a few works concerning \acp{NCA} for image synthesis or enhancement, which we review below.

\noindent\textbf{Growing images from a seed:} Early works, including the initial NCA paper and demo by \citet{mordvintsev2020growinga}, focus on generating small-scale emoji images.
This method aims to understand biological growth, rather than generating a diverse set of images. The \ac{NCA} must be retrained for each image, hindering its applications to real-world problems.
Following the work of~\citet{randazzo2021adversarial} changes the behavior of a trained \ac{NCA} using a single symmetric matrix $A$, which is multiplied to each state after each \ac{NCA} step. In the experiments, they managed to reprogram the \ac{NCA} into removing several parts of the generated lizards' limbs or changing its color, all while maintaining a mostly stable pattern. 
Subsequently, \citet{mordvintsev2022growing} and \citet{randazzo2023growing} propose \ac{NCA} architectures that encode isotropic (rotationally equivariant) rules that, hence, can generate the same image in rotated variants. The specific rotation is either determined by placing multiple seeds or inferred randomly due to the stochastic nature of the \ac{NCA}.

\noindent\textbf{Growing Graphs:} Several \acp{NCA} have been extended to graphs. Some of which have been evaluated by learning to grow a fixed graph structure~\citep{dwyer2023training,gala2023enequivariant,grattarola2021learning}.

\noindent\textbf{GAN-NCA-hybrids:} The first work that uses \ac{NCA} in a generative adversarial manner is GANCA~\citep{otte2021generativea}, which replaces the generator of a vanilla GAN with an \ac{NCA} model.
The generator is conditioned on the edges of emojis and generates fully colored emojis.
For better performance and training stability, the authors employ modifications akin to the tweaks introduced in \ac{WGAN}~\citep{arjovsky2017wassersteina}.
StyleGANCA~\citep{luu2026coarse} extends the idea with a multi-resolution strategy~\citep{lemke2025octreenca} and style-conditioning~\citep{karras2020analyzing}, resulting in an algorithm for conditional and unconditional medical image generation.

\noindent\textbf{VAE-NCA-hybrids:} \citet{palm2022variationala} implement an \ac{NCA} in a variational autoencoder framework.
The encoder is a standard CNN-based variational encoder.
The decoder is a single \ac{NCA} that is iteratively applied for 8 steps after doubling the resolution using an unpooling operation.
After 5 upscaling operations, the \ac{NCA} produces the final output image.
Initially, the VAE generates poorly looking images when sampling from the prior.
This issue can be fixed by introducing the beta-VAE~\citep{higgins2017betavae} loss ($\beta=100$) and averaging multiple decoder runs from the same latent code.
While the generated images look better visually, the averaging removes any detail.

\noindent\textbf{Diffusion NCA:} By far, most interest in generative modeling with \acp{NCA} has been in diffusion models, which is probably due to their great success in different fields in recent times~\citep{dhariwal2021diffusion}.
The first published diffusion \ac{NCA} article~\citep{elbatel2024organisma} utilizes the ViTCA~\citep{tesfaldet2022attentionbased} architecture performing latent diffusion~\citep{rombach2022highresolution}.
The architecture includes adaptive LayerNorm~\citep{perez2018film} inspired by the Diffusion Transformer architecture~\citep{peebles2023scalable}.
FourierDiff-NCA~\citep{kalkhof2024frequencytime} relies on the typical fully connected architecture of \acp{NCA}, but performs diffusion in the Fourier space, which proves to be more efficient than diffusion in the image space directly.

\noindent\textbf{Direct synthesis through NCAs:}
TeNCA~\citep{lang2025temporal} is a single \ac{NCA} for generating videos of the temporal dispersion of contrast agent in breast \acp{MRI}.
The data consists of paired images before and after applying the contrast agent with different delays after the contrast agent was injected.
The authors apply the loss at certain time steps for which a post-contrast image exists.
The trained model can synthesize images, resulting in a video showing the smooth temporal spread of contrast agent in the breast \ac{MRI}.

\noindent\textbf{Texture Synthesis:}
\citet{mordvintsev2021texturea} are the first to propose the use of \acp{NCA} for texture synthesis.
They adapt the original NCA training approach \citep{mordvintsev2020growinga} by introducing a perception-based texture loss, using a  pretrained VGG model to extract high-level appearance features.
Further research on NCA-based texture generation broadly falls into two categories: Methods that explore architectural modifications to improve texture generation abilities, or works that extend NCA-based texture generation to new domains or applications.
We will start by discussing papers that primarily propose architectural modifications:\\ 
As follow-up work to their original paper (\citet{mordvintsev2020growinga}), \citep{mordvintsev2021mnca} develop $\mu$NCA, a family of texture generation NCAs with as few as 68 total parameters.
$\mu$NCA introduces an optimal-transport-based loss function, as well as several modifications to the standard NCA architecture.
To achieve the reduction in model size, $\mu$NCA removes the latent linear layer in the update function and adapts the perception to apply the perception filters to alternating channels, reducing the size of the perception vector.
\citet{petersen2023exploring} propose a variant of the \ac{NCA} architecture that utilizes multiscale perception. \\ 
Due to their large memory requirements during training, generating high-resolution textures remains a challenging task. Possible solutions to this problem are developed by \citet{pajouheshgar2025neural} and \citet{pajouheshgar2024noisenca}.
\citet{pajouheshgar2025neural} propose the use of an \ac{NCA} operating in a coarse latent space combined with a lightweight decoder network to enable high-resolution texture generation.
\citet{pajouheshgar2024noisenca} demonstrate that using an initial noisy seed instead of stochastic cell-state updates results in more robust behavior with respect to the discretization in both space and time.
This enables upscaling of the perception range of the cell-grid at test-time, resulting in higher resolution textures.
Similarly, the approach of \citep{pajouheshgar2024noisenca} also gives the user control over the speed at which textures evolve.\\
\citet{catrina2024multitexture} develop an \ac{NCA} model capable of learning multiple textures by encoding an identifier for the target texture in the initial seed. Training a single \ac{NCA} on multiple target textures improves parameter efficiency and allows for combining multiple textures in a single cell grid as well as smooth interpolation between textures.\\ 
Several works aim to extend NCA-based texture generation beyond the setting of static 2D textures. \citep{pajouheshgar2024mesh} propose the Mesh Neural Cellular Automata, a variant of the \ac{NCA} architecture that uses undirected graphs as the state space, allowing for efficient texture synthesis directly on 3D surface meshes.
The authors explore a broad range of applications, including text-guided synthesis, dynamic texture synthesis, and training multiple compatible Mesh-\acp{NCA} to allow for smooth interpolation between different textures on the same surface mesh.\\ 
\citet{wang2025volumetrica} choose a voxel-based approach to 3D generation.
They introduce the Volumetric Neural Cellular Automata (VNCA), focusing on stylized smoke rendering as an application.
During training, VNCA is conditioned on both a density field and a motion field, allowing it to generalize to new shapes and dynamics at inference time.\\
\citet{pajouheshgar2023dynca} use \acp{NCA} to generate dynamic video textures.
To adapt NCA-based texture generation to video, they introduce a novel optical-flow-based motion loss, among other modifications.
Their approach generalizes to videos of arbitrary length and size. 
\citet{seibert2024reconstructinga} use \acp{NCA} to generate realistic surface structures for a wide range of materials.
Their main innovation is the use of statistical descriptors of the surface structure for the loss.

\subsection{Image Classification}
The first paper to use \acp{NCA} for image classification was a follow-up work to the initial \ac{NCA} paper by \citet{mordvintsev2020growinga}, proposing \textit{Self-classifying MNIST digits}~\citep{randazzo2020selfclassifying}.
This approach leverages designated output channels to store the final prediction, adding to the hidden channels and fixed input image channel(s).
After running the \ac{NCA} for 200 time steps, the output channels are populated such that a channel-wise argmax yields the prediction for each image pixel.
In the case of the MNIST dataset, this approach leads to interesting self-organizing behavior, especially in out-of-distribution cases like a chimera of two different digits.

A more recent, practical application in medical imaging is an approach for blood cell classification, which is proposed by \citet{deutges2024neurala}.
This approach leverages an \ac{NCA} model that is applied to blood cell images for 64 steps.
Afterwards, the channel-wise maximum is taken to obtain a single low-dimensional latent vector, which is then fed to a 2-layer neural network to obtain a prediction vector.
This idea leverages the hidden channels for the final prediction, instead of appending channels for each category, which would become costly for a multi-class problem.

Several works have explored \ac{NCA} as a robust feature extractor for classification.
\citet{ihm2024localized} use a pre-trained \ac{NCA} for robust feature extraction combined with a ResNet for classification.
\citet{tesfaldet2022attentionbased} pre-train an NCA using a masked autoencoding objective and extract class labels using linear probing.

\subsection{Handling Distribution Shifts}
In medical image analysis, models often encounter distribution shifts, for example, due to changes in the patient demographics or the scanning protocol.
In those cases, we desire mechanisms in the model to automatically deal with those distribution shifts to detect or reject them before giving a misleading prediction.
\citet{korevaar2024generalization} do not propose a methodological contribution, but they investigate the generalization of \acp{NCA} to \ac{OOD} data, which was not seen during training.
They construct a scenario of retinal \ac{OCT} images and cardiac MRIs obtained with scanners from different vendors, creating a realistic scenario.
On the retinal data, the \ac{NCA} outperformed the baseline UNet on \ac{ID} and \ac{OOD} data.
On the cardiac MRIs, the \ac{NCA} performed slightly worse than the UNet on ID data, but significantly better than the UNet on \ac{OOD} data.
Concluding, the study of~\citet{korevaar2024generalization} suggests that \acp{NCA} might be better suited for out-of-the-box domain generalization than UNets.

While uncertainty estimation and \ac{OOD} detection are established practices in the analysis of convolutional- and transformer-based models for image segmentation~\citep{gonzalez2021detecting,gonzalez2022distancebased,lemke2024distributionaware,fuchs2022practical} and classification~\citep{zhang2021out,graham2023denoising}, there are only a few works discussing uncertainty estimation of \acp{NCA} or proposing tailored \ac{OOD} detection mechanisms.
Generally, \acp{NCA} have been shown to generalize better to unseen data containing imaging artifacts or acquired at different resolutions.
However, OOD detection is a critical task, especially for medical image analysis~\citep{yang2024generalized}.
In the following, we present the only two methods for \ac{OOD} detection during inference time.
To the time of writing, we were not able to find any other works that propose uncertainty estimation or \ac{OOD} detection techniques for \acp{NCA}.

\noindent\textbf{NQM:} \citet{kalkhof2023m3dnca} propose the \ac{NQM} to assess prediction uncertainty of \acp{NCA}.
The non-deterministic inference algorithm produces different segmentation masks for each random seed.
Running the inference multiple times and computing the standard deviation delivers an estimate of the model's uncertainty.
An uncertainty threshold can successfully detect OOD samples at inference time.

\noindent\textbf{LANCA:} \citet{ihm2024localized} propose LANCA, a pipeline for training robust, \ac{NCA}-based feature extractors for image processing.
Using an Autoencoder architecture, LANCA trains an encoder \ac{NCA} to extract local image features.
Each cell learns to encode its environment in the hidden state, resulting in a fully localized approach to feature extraction.
As demonstrated by \citet{ihm2024localized}, a two-stage image classification pipeline consisting of a LANCA encoder and a shallow ResNet displays improved robustness across a wide range of image corruptions compared to a ResNet of equal size without the LANCA encoder.

In the following, we will review methods that work on adaptation to the shifted distribution. 

\noindent\textbf{Leveraging distribution shifts for continual learning:} NCAdapt~\citep{ranem2025ncadapta} is a continual learning algorithm for \acp{NCA} based on uncertainty-aware model expansion.
Continual learning deals with shifting distributions by training on a new dataset of the shifted distribution without access to the previous distribution.
During training, the performance on the data of the earlier distributions should be maintained, increasing and expanding the knowledge of the model~\citep{chen2017lifelong}.
Unfortunately, continual learning algorithms over-adapt on new data, losing knowledge on solving the earlier tasks, which is known as \textit{catastrophic forgetting}~\citep{french1999catastrophic}.
NCAdapt circumvents this issue for hippocampus segmentation by introducing a few (384) domain-specific parameters into the \ac{NCA} architecture.
While parameter-expanding continual learning algorithms are nothing novel~\citet{rusu2022progressive}, NCAdapt is probably one of the methods requiring the fewest domain-specific parameters, rendering the model scalable even for large sequences of domain shifts.
The domain-specific parameters are inferred by running multiple inference passes, each of which uses a separate set of domain-specific parameters.
The domain identity is inferred by the minimum \ac{NQM}~\citep{kalkhof2023m3dnca} score of each previous run. 

\noindent\textbf{Unsupervised Fine-tuning:} If, after a data shift, new images are recorded but not annotated, one has to use an unsupervised algorithm to adapt to the new distribution.
\citet{kalkhof2024unsupervised} propose an algorithm for unsupervised fine-tuning for \acp{NCA}, so the model will adapt to the new distribution without requiring any labels of the novel data.
The method makes use of the \ac{NQM}~\citep{kalkhof2023m3dnca} metric that measures the quality of the prediction based on the output variance of multiple \ac{NCA} runs.
By reducing the standard deviation and improving the \ac{NQM}, the model will automatically adapt to the new distribution in an unsupervised manner.

\subsection{Data Compression}

With the ongoing digitization of the healthcare sector, the demand for medical data records on disk space has become a challenge.
In 2013, roughly 153 Exabytes of medical data were produced.
It was projected that in 2020, 2314 Exabytes of medical data would be produced, which corresponds to a yearly growth of $48\%$~\citep{stanfordmedicine2017harnessing}.
\citet{choi2023current} report detailed information of the picture archiving and communication systems storage requirements at the Seoul National University Hospital from 2015 to 2022.
They estimated a yearly growth of $11.25\%$, reaching up to $\SI{3000}{\tera\byte}$ of data in 2029.
Such enormous data generation puts high storage demands on clinics, which are typically not well-equipped.
To alleviate the space requirements, compression schemes for medical image data have already been discussed~\citep{tong2025survey,xin2021lossless}.
The same issues appear in remote sensing with satellites and drones taking high-resolution scans of the Earth. NASA reports to have more than $\SI{128}{\peta\byte}$ of storage with a daily growth of $\SI{147}{\tera\byte}$~\citep{earthsciencedatasystems2024esds}.
A similar issue arises from genomic sequencing data. The genome of a single human can reach up to $\SI{100}{\giga\byte}$ of data. \citet{greenfield2019importance} expect the data demand for genome sequences to reach $\SI{40}{\exa\byte}$ per year by 2025.

\citet{falcao2025unraveling} propose to use \acp{NCA} for natural image compression.
The method splits the image into smaller patches and trains a separate growing NCA~\citep{mordvintsev2020growinga} for each patch.
The parameters of each \ac{NCA} can be represented more efficiently than the pixels 
themselves.
Although this is so far the only work investigating image compression with \ac{NCA}, with an increase in other neural methods for data compression, we believe that this so far under-explored area could soon gain more traction.

\subsection{Simulation}
\acp{CA} have long been used to model \mbox{spatio-temporal} processes governed by local interactions.
Since \acp{PDE} can, in principle, be reformulated as \acp{CA} \citep{yang2010cellular}, \acp{NCA} naturally extend this approach to learn such dynamics directly from data.
\citet{li2001calibrationa,li2002neuralnetworkbased,yeh2003simulation,okwuashi2012gis} train a neural network that encodes the update algorithm for a single \ac{CA} step.
Consequently, \acp{NCA} have been applied to urban development, land-use transitions, microstructure evolution, atmospheric dispersion, and biological pattern formation.
Because the training is non-iterative, their formulation does not fit our \ac{NCA} definition; regardless, we have decided to include them in this work due to the large community behind it.

\noindent\textbf{Urban development and land-use simulation:}
Urban development and land-use change are spatial processes emerging from local transitions such as densification or infrastructure expansion.
Modeling these dynamics requires combining neighborhood interactions with contextual features.
Early work integrated neural networks with \acp{CA}, where the network learns the local update rule and is embedded into the \ac{CA} for iterative prediction.
These models operate on hand-engineered features such as proximity to infrastructure, terrain properties, and neighborhood statistics. Representative approaches include \citet{li2001calibrationa}, \citet{li2002neuralnetworkbased}, \citet{yeh2003simulation}, and \citet{okwuashi2012gis}.
\citet{almeida2008using} applied similar methods for intra-urban land-use modeling.
Later extensions combined neural-network-based \acp{CA} with Markov chains \citet{xu2019simulation}, while \citet{xing2020novel} proposed a non-Markovian deep learning formulation.
More recently, fully neural \ac{CA} models have demonstrated high accuracy in large-scale urban and forest simulations \citep{zhang2024neural}.

\noindent\textbf{Material and microstructure simulation:}
In materials science, \acp{NCA} have been applied to processes traditionally modeled by \acp{PDE}, such as solidification and grain growth. \citet{tang2023neural} trained a convolutional \ac{NCA} to reproduce microstructure evolution, achieving competitive accuracy while being orders of magnitude faster than classical \ac{CA} methods.

\noindent\textbf{Biological pattern formation and \ac{PDE}-driven dynamics:}
\citet{richardson2024learning} trained \acp{NCA} following \citet{mordvintsev2020growinga} to reproduce \ac{PDE}-driven pattern formation, such as Turing patterns, while preserving key symmetries. This demonstrates that \acp{NCA} can serve as compact learned approximators of structured physical dynamics.
\citet{navarin2024physicsinformed} apply an \ac{NCA} on a graph for compartmental modelling of \mbox{Covid-19}.

\noindent\textbf{Other physical simulations:}
\ac{ANN-CA} hybrids have also been used for dispersion modeling. \citet{lauret2016atmospheric} combined \acp{CA} with neural networks to predict atmospheric methane (CH\textsubscript{4}) dispersion, approximating complex computational fluid dynamics systems with local learned rules.

Overall, \acp{NCA} provide an efficient framework for modeling systems where global behavior emerges from repeated local interactions, enabling data-driven simulation of complex spatio-temporal processes.

\section{Potential and Limitations}\label{sec:discussion}

In the following, we discuss the value of \acp{NCA} for various applications, the potential of recent developments, and the technology's potential challenges and research gaps.

\subsection{Resource Efficiency}

As our world is already significantly impacted by climate change, the trend of energy- and resource-hungry AI models is alarming and needs to be counteracted.
Increasingly large models currently dominate the \ac{AI} ecosystem, requiring powerful and new hardware, and placing high demands on power consumption.
\acp{NCA} are on the other side of the spectrum, showing the potential to be well-suited for resource-friendly and sustainable \ac{AI}.
Further, they can run locally on small-scale ubiquitous hardware platforms, performing challenging tasks like medical image segmentation in a reasonable time at sufficient accuracy.

Several works have shown that \acp{NCA} are capable of running on small-scale hardware such as smartphones~\citep{kalkhof2024unsupervised}, single board computers like the Raspberry Pi~\citep{kalkhof2023mednca,kalkhof2023m3dnca,ranem2024ncamorph}, and even microcontrollers~\citep{krumb2025encapsulatea}.
Even training is possible on smartphones, for instance, in federated setups as described by \citet{kalkhof2024unsupervised} and \citet{lemke2025equitablea}.
In such a federated setup, the resource efficiency has interesting side effects, making it possible to efficiently rely on homomorphic encryption to preserve patient privacy in multi-client training~\citep{lemke2025equitablea}, which is hardly possible with \ac{CNN} or \ac{ViT}-style architectures due to the encryption's poor runtime scaling with respect to the number of parameters.

Despite the fast and cheap inference, a clear limitation of \acp{NCA} is currently their high demand to \ac{VRAM} on the \ac{GPU} during training.
Training these models employs \ac{BPTT}, which is similar to the fashion in which Recurrent Neural Networks (RNN) are trained~\citep{mordvintsev2020growinga}.
Since all \ac{NCA} timesteps need to be unrolled for backpropagation, $k$ gradient updates need to be stored in \ac{VRAM} during training, which is costly.
Hence, training an \ac{NCA} on consumer-grade \acp{GPU} requires considerations regarding the maximum image size, number of hidden channels, or the number of time steps.

Recent works propose improvements to the vanilla \ac{NCA} training process to alleviate the high \ac{VRAM} requirements.
For instance, M3D-NCA by \citet{kalkhof2023m3dnca} uses a combination of patchification and scaling to allow for efficient training of a 3D \ac{NCA}.
\citet{lemke2025octreenca} propose a cascaded \ac{NCA} that operates at multiple image scales in a sequence.
A convenient side effect of multi-scale \ac{NCA} pipelines is that they manage to capture long-range dependencies as well as local features, which is desirable for various applications that require global knowledge to be incorporated in the model.

\subsection{Generalization Capabilities}
Unlike \ac{CNN}- or \ac{ViT}-based architectures, \acp{NCA} are inherently invariant to the input image size during training and inference~\citep{kalkhof2023mednca}.
\acp{CNN} often require a fixed image size, or are prepended by an adaptive pooling layer.
In any case, image information needs to be discarded if a \ac{CNN} is fed a larger image during inference, while \acp{NCA} make use of all pixels.
For \ac{ViT}, extensions like VarViT exist to deal with variable input image sizes~\citep{varma2024varivit}, but are not yet established as a default architectural change.

Further, it was shown that \acp{NCA} are translation invariant~\citep{kalkhof2023mednca}.
Translation invariance is a naturally useful property for models trained on curated medical image datasets, which are centered on the anatomy of interest.
However, it cannot always be assumed that the user centers each image to the region of interest, leading to silent failure of \ac{CNN}-based models.
As \acp{CNN} directly encode global features, and \ac{ViT} uses a positional encoding, these models fail in such a scenario.
\acp{NCA} on the other hand, operate locally and propagate their updates over time, with each cell executing the same program, so that they are mostly invariant to global translation or cropping.
Generalization in terms of domain generalization is examined in two works that attempt to solve ARC AGI tasks with \ac{NCA}~\citep{xu2025neural,korevaar2024generalization}.

\subsection{Future challenges and opportunities}
Our analysis highlights several challenges with \acp{NCA} and its literature.
We believe that exploring solutions to those challenges opens interesting research avenues in the \ac{NCA} landscape.

First and foremost, the under-exploration of real world applications of \acp{NCA} is still a limitation.
Many theoretical advancements are demonstrated on oversimplified tasks, such as emoji datasets, benchmark datasets such as \mbox{CIFAR-10}, \mbox{(Fashion-)MNIST}, or artificially created data, which do not yet translate to actual real-world or medical images.
Only a small subset of architectural variations is evaluated on challenging real-world data, such as \ac{CT} or \ac{MRI} images -- compared to well-studied architectures based on \ac{CNN} or \ac{ViT}.
This gap can be overcome by fostering the integration of \acp{NCA} into easy-to-use general-purpose software frameworks, similar to the idea of \textit{nnUNet}~\citep{isensee2021nnunet}.
OctreeNCA by~\citet{lemke2025octreenca} is an initial effort to integrate \acp{NCA} into nnUNet, which is part of the Lifelong nnUNet code repository~\citep{gonzalez2023lifelong}.

From a technical perspective, the main issue with \acp{NCA} is their slow and resource-heavy training.
Due to the \ac{BPTT}, \ac{NCA} training can be expensive in terms of energy consumption and \ac{VRAM} demand.
As \acp{NCA} can only communicate locally, they transfer information for a single pixel at a time, which demands many \ac{NCA} steps for high-resolution data, which further exacerbates the already demanding training.
Works that attempt to solve this issue often resort to domain-specific adaptations, which have not yet been shown to generalize to multiple applications of \acp{NCA} in medical imaging.

The expensive training, in turn, limits the scalability of \acp{NCA} to more complex tasks, such as the segmentation of multiple classes.
Most \ac{NCA}-based classification (or segmentation as pixel-wise classification) methods focus on predicting a single class~\citep{kalkhof2023mednca, kalkhof2023m3dnca}, only a few methods predict multiple classes~\citet{lemke2025octreenca}.
To the best of our knowledge, there are no \ac{NCA} algorithms capable of segmenting more than $10$ classes at once.

The \ac{NCA}'s small number of parameters limits its application to complex tasks. 
Hence, most \acp{NCA} are trained for binary segmentation of a single class~\citep{kalkhof2023mednca}, classifying less than 13 classes in a constrained setting~\citep{deutges2024neurala}, or generating a single texture~\citep{pajouheshgar2023dynca}.
Only few works have trained \acp{NCA} for solving complex tasks, which usually results in a serious performance degradation compared to the confined scenario~\cite{lemke2026sterilizable}.

While the AI research community has established an intuition on the operating principles of \acp{CNN} and Transformers, such understanding is still missing for \acp{NCA}.
Early works explore the information propagation and state space evaluation of \acp{NCA}~\citep{stovold2026visualising,kvalsund2026stability}; however, there is no general understanding of how \acp{NCA} solve tasks, be it image segmentation or emoji growing.

Since \acp{NCA} rely solely on local information exchange with direct neighbors, \acp{NCA} naturally experience a global information bottleneck.
Information is diffused slowly through multiple NCA iterations, leading to a global communication bottleneck.
Recent works have found various ways to ease global knowledge accumulation~\citep{lemke2025octreenca,kalkhof2024frequencytime,pio-lopez2026brainca,yang2025attention}, but no general-purpose solution exists.

On the upside, \acp{NCA} require lower resource consumption during inference time. 
Also, the low parameter count of \acp{NCA} allows deployment on small-scale hardware, where integration of AI accelerators is impossible, especially in recent times of hardware shortages.
Adding to the small size of \ac{NCA} models themselves, \acp{NCA} could be leveraged for medical image compression -- however, this area of \ac{NCA} research needs further investigation.
Finally, multiple works have discovered the increased robustness of \acp{NCA} on noisy or shifted data.
This can likely be attributed to the unique emergent properties of \acp{NCA}.

\section{Conclusion}\label{sec:conclusion}

In this paper, we conducted a review of the current state of the art of \acp{NCA}, with a focus on their application in imaging tasks.
We have found several works exploring the applicability of these exciting new models in various downstream tasks, including semantic segmentation, image registration, and texture generation.
While \acp{NCA} have interesting properties such as their compellingly small model size, there are still several challenges that are yet to be overcome, such as their high demands on \ac{VRAM} during training.
Further, \acp{NCA} do not yet seem to be fully understood regarding their explainability, nor the uncertainty of their predictions.
Initial attempts in this direction look promising, but more work is required towards more robust evaluations.
With this review, we hope to provide researchers with access to the exciting field of \acp{NCA}, hoping for the field to gain more traction in the future in order to exchange knowledge between research communities towards more efficient and robust architectures.

\noindent\textbf{Acknowledgements:} Authors DML and JAS received funding from HELMHOLTZ IMAGING, a platform of the Helmholtz Information and Data Science Incubator.

\bibliography{bibliography}

\end{document}